\pdfoutput=1

\documentclass[11pt]{article}

\usepackage[]{emnlp2022}

\usepackage{times}
\usepackage{latexsym}

\usepackage[T1]{fontenc}

\usepackage[utf8]{inputenc}

\usepackage{microtype}
\usepackage{inconsolata}
\usepackage{graphicx}

\usepackage{amsmath}
\usepackage{amsfonts}
\usepackage{multicol}  
\usepackage{multirow}
\usepackage{booktabs}
\usepackage{colortbl}
\usepackage{xspace}
\usepackage{lato}
\usepackage{algorithm} 
\usepackage{algorithmicx} 
\usepackage{algpseudocode}
\usepackage{radarchart}

\newcommand{\name}{\textsc{STaR}\xspace}
\newcommand{\spider}{\textsc{Spider}\xspace}
\newcommand{\sparc}{\textsc{SParC}\xspace}
\newcommand{\cosql}{\textsc{CoSQL}\xspace}
\newcommand{\bert}{\textsc{Bert}\xspace}
\newcommand{\roberta}{\textsc{Roberta}\xspace}
\newcommand{\grappa}{\textsc{Grappa}\xspace}
\newcommand{\score}{\textsc{SCoRe}\xspace}
\newcommand{\electra}{\textsc{ELECTRA}\xspace}
\newcommand{\picard}{\textsc{Picard}\xspace}
\newcommand{\bart}{\textsc{Bart}\xspace}

\def\eg{\emph{e.g.}}
\def\ie{\emph{i.e. }}

\usepackage{stfloats}
\usepackage{adjustbox}

\makeatletter
\g@addto@macro\normalsize{%
  \abovedisplayskip 1pt plus 1pt minus 1pt%
  \belowdisplayskip \abovedisplayskip
  \abovedisplayshortskip 2pt plus1pt  minus1pt%
  \belowdisplayshortskip 2pt plus1pt minus1pt%
}
\g@addto@macro\small{%
  \abovedisplayskip 2pt plus 1pt minus 1pt%
  \belowdisplayskip \abovedisplayskip
  \abovedisplayshortskip 2pt plus1pt  minus1pt%
  \belowdisplayshortskip 2pt plus1pt minus1pt%
}
\g@addto@macro\footnotesize{%
  \abovedisplayskip 2pt plus 1pt minus 1pt%
  \belowdisplayskip \abovedisplayskip
  \abovedisplayshortskip 2pt plus1pt  minus1pt%
  \belowdisplayshortskip 2pt plus1pt minus1pt%
}
\makeatother

\makeatletter
\renewcommand{\maketag@@@}[1]{\hbox{\m@th\normalsize\normalfont#1}}
\makeatother
\title{\name: SQL Guided Pre-Training for Context-dependent\\ Text-to-SQL Parsing
}

\author{Zefeng Cai$^{1, 2, \includegraphics[scale=0.02]{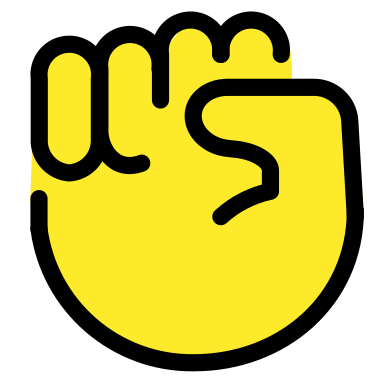}}$, Xiangyu Li$^{1, 2, \includegraphics[scale=0.02]{first.png}}$, Binyuan Hui$^{3}$, Min Yang$^{2}$\footnotemark[2], Bowen Li$^{3}$, \\
\bf{Binhua Li$^{3}$, Zheng Cao$^{3}$, Weijie Li$^{3}$, Fei Huang$^{3}$, Luo Si$^{3}$, Yongbin Li$^{3}$\footnotemark[2]}\\
$^1$ University of Science and Technology of China \\
$^2$ Shenzhen Institute of Advanced Technology, Chinese Academy of Sciences \\
$^3$ DAMO Academy, Alibaba Group \\
\texttt{\{zf.cai, xy.li3, min.yang\}@siat.ac.cn} \\
\texttt{\{binyuan.hby, binhua.lbh, shuide.lyb\}@alibaba-inc.com}\\
}

\begin{document}
\maketitle
\renewcommand{\thefootnote}{\fnsymbol{footnote}}
\footnotetext{\includegraphics[scale=0.02]{first.png} Equal contribution.}
\footnotetext[2]{Corresponding authors.}
\begin{abstract}
In this paper, we propose a novel SQL guided pre-training framework \name for context-dependent text-to-SQL parsing, which leverages contextual information to enrich natural language (NL) utterance and table schema representations for text-to-SQL conversations. Concretely, we propose two novel pre-training objectives which respectively explore the context-dependent interactions of NL utterances and SQL queries within each text-to-SQL conversation: (i) schema state tracking (SST) objective that tracks and explores the schema states of context-dependent SQL queries in the form of schema-states by predicting and updating the value of each schema slot during interaction; (ii) utterance dependency tracking (UDT) objective that employs weighted contrastive learning to pull together two semantically similar NL utterances and push away the representations of semantically dissimilar NL utterances within each conversation. In addition, we construct a high-quality large-scale context-dependent text-to-SQL conversation corpus to pre-train \name. Extensive experiments show that \name achieves new state-of-the-art performance on two downstream benchmarks (\sparc and \cosql), significantly outperforming previous pre-training methods and ranking first on the leaderboard. We believe the release of the constructed corpus, codebase and pre-trained \name checkpoints would push forward the research in this area. For reproducibility, we release our code and data at \url{https://github.com/AlibabaResearch/DAMO-ConvAI/tree/main/star}.
\end{abstract}

\section{Introduction}
Text-to-SQL parsing \cite{wikisql,yu2018spider,wang2022proton,qin2022sun} aims to translate natural language (NL) questions into executable SQL queries, which enables the users who are unfamiliar with SQL to query databases with natural language.
Pre-trained language models (PLMs) have proved to be powerful in enhancing text-to-SQL parsing and yield impressive performances, which benefit from the rich linguistic knowledge in large-scale corpora. However, as revealed in previous works \cite{yin2020tabert,yu2021grappa,Qin2022ASO}, there are intrinsic discrepancy between the distributions of tables and plain texts, leading to sub-optimal performances of general PLMs such as \bert \cite{devlin2018bert}, \roberta \cite{liu2019roberta}, \electra \cite{clark2020electra}.
Recently, some studies \cite{yu2021grappa,yu2020score,Shi2020LearningCR,Deng_2021,DBLP:journals/corr/abs-2107-07653,liu2021awakening} alleviate the above limitation by designing tailored tabular language models (TaLMs) for text-to-SQL parsing, which simultaneously encode NL questions and tables.

\begin{figure}
    \centering
    \includegraphics[width = 7.5cm]{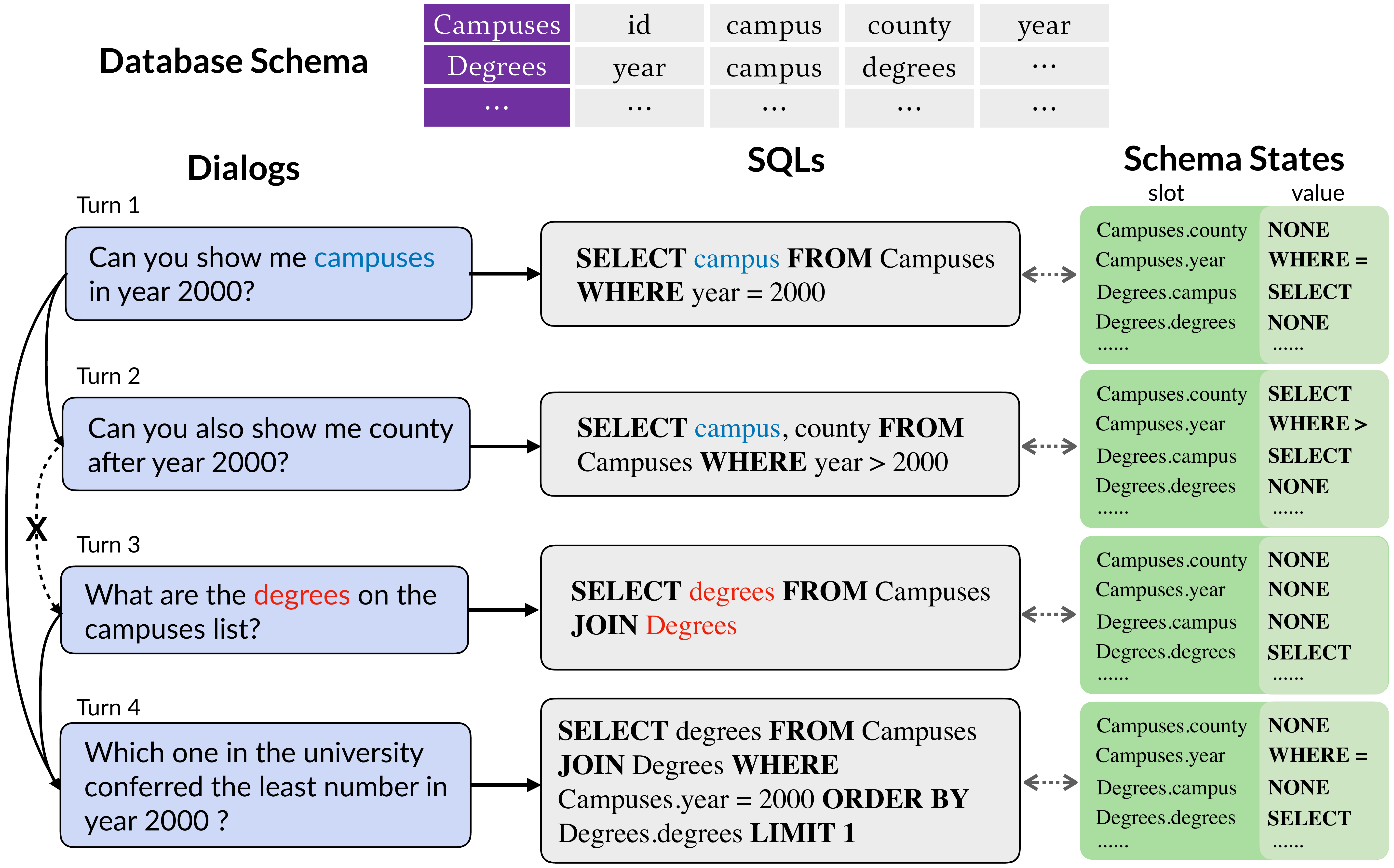}
    \caption{An example of cross-domain context-dependent Text-to-SQL conversation. Here, each database schema refers to the table/column names of databases and each schema state refers to a slot-value pair, whose slot is a column/table name (e.g., Degrees.campus) and its value is a SQL keyword (e.g., SELECT). ``x'' indicates that the semantic/intent is switched between Turn2 and Turn3 utterances.}
    \label{fig:cases}
\end{figure}

Despite the remarkable progress of previous TaLMs, they still suffer from technical challenges in the context-dependent setting. 
\textbf{First}, existing TaLMs merely explore contextual information to
enrich utterance representations without considering the interaction states determined by history SQL queries, which are relevant to the user intent of current utterance. Nevertheless, the trace and usage of historical SQL information can contribute greatly to model the current SQL query, as SQL conveys user intent in a compact and precise manner.
As shown in Figure \ref{fig:cases}, the second SQL query is more likely to select the contents from the ``Compuses'' table since  the first SQL query mentioned that table. Although tracking schema states is essential to keep track of user requests for context-dependent text-to-SQL parsing, how to model, track and utilize schema states throughout a conversation has not yet been explored in previous TaLMs.
\textbf{Second}, context-dependent text-to-SQL parsing needs to effectively process context information so as to help the system better parse current NL utterance, since users may omit previously mentioned entities as well as constraints and introduce substitutions to what has already been stated. Taking Figure  \ref{fig:cases} as an example, the second utterance omit the implicit constraint of ``campuses in year 2000'' as mentioned in the first utterance. However, most prior TaLMs primarily model stand-alone NL utterances without considering the context-dependent interactions, which result in sub-optimal performance. 
Although \textsc{SCoRe} \cite{yu2020score} model the turn contextual switch by predicting the context switch label between two consecutive user utterances, it ignores the complex interactions of context utterances and cannot track the dependence between distant utterances. For instance, in Figure \ref{fig:cases}, \score fails to capture the long term dependency between the first and the fourth utterances since there is a switch between the second and the third utterances. 

In this paper, we propose a novel pre-training framework \name for context-dependent text-to-SQL parsing, which explores the multi-turn interactions of NL utterances and SQL queries within each conversation, respectively. 
\textbf{First}, we propose a schema state tracking (SST) objective to keep track of SQL queries in the form of schema-states, which predicts the value (a SQL keyword) of each schema slot of the current SQL query given the schema-state representation of previously predicted SQL query. By introducing the schema-states to represent SQL queries, we can better capture the alignment between the the historical and current SQL queries, especially for the long and complex SQL queries. 
\textbf{Second}, we propose an utterance dependency tracking (UDT) objective to capture complex semantic dependency of sequential NL questions, which employs weighted contrastive learning to pull together semantically similar NL utterances and push away dissimilar NL utterances within each conversation. 
A key insight is that the utterance corresponding to similar SQL will be more semantically relevant, as SQL is a highly structured indication of user intent.
Concretely, we propose two novel similarity functions (SQL semantic similarity and SQL structure similarity) to comprehensively construct appropriate positive and negative NL question pairs. 

We summarize our main contributions as follows. 
(1) To the best of our knowledge, we are the first to propose a schema state tracking (SST) objective for context-dependent TaLM, which tracks and updates the schema states of the context-dependent SQL queries in the form of schema states. 
(2) We propose an utterance dependency tracking (UDT) objective to capture complex semantic information of sequential NL questions, which employs weighted contrastive learning with two novel SQL-oriented similarity functions to pull together two semantically similar NL utterances and push away the representations of dissimilar NL utterances within each conversation. 
(3) We construct a high-quality large-scale context-dependent text-to-SQL conversation corpus to pre-train \name. Experiments show that \name achieves new state-of-the-art performance on two downstream benchmarks (\sparc and \cosql) and ranking first on the leaderboard.

\begin{figure*}
    \centering
    \includegraphics[width = 16cm]{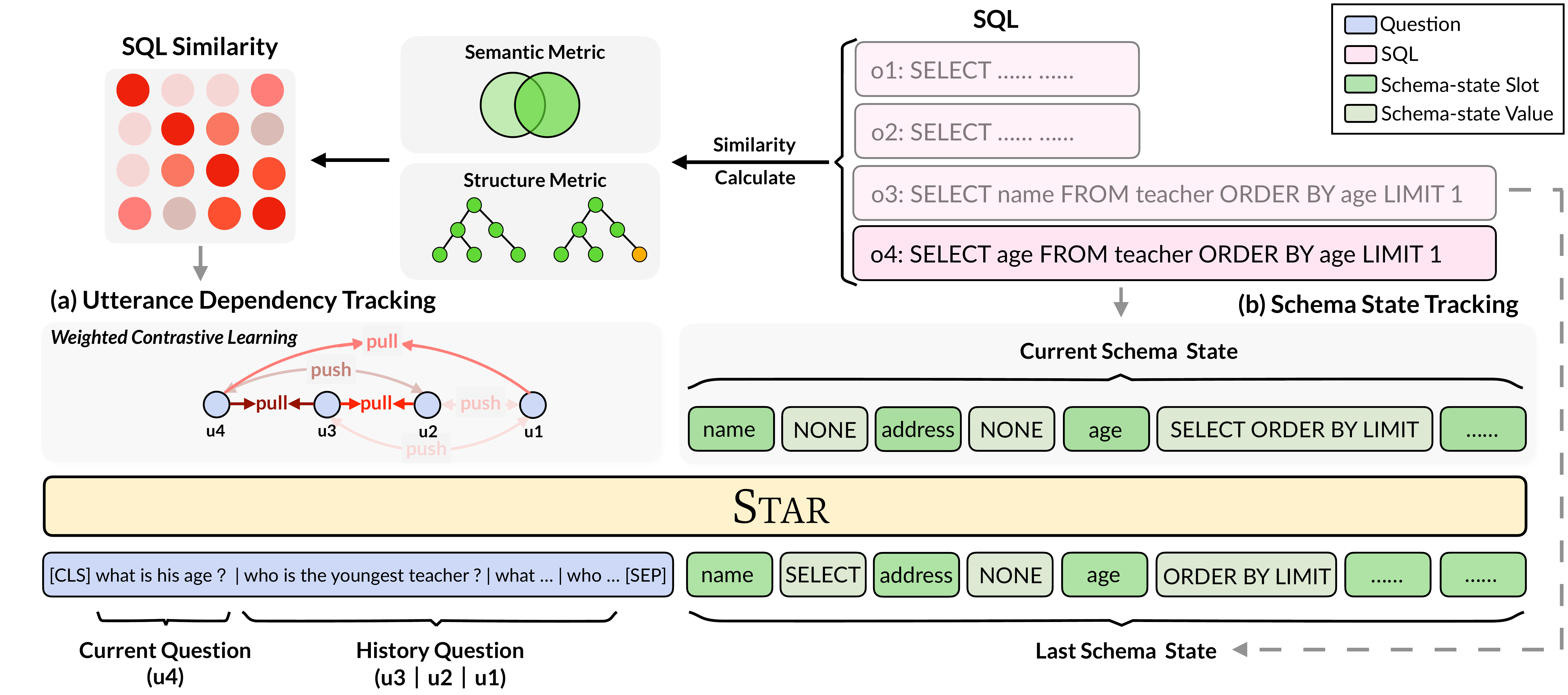}
    \caption{The overview of the proposed \name framework consisting of two novel pre-training objectives: (a) the utterance dependency tracking and (b) the schema state tracking. For brevity, we do not show the masked language modeling objective here. }
    \label{fig:overview}
\end{figure*}

\section{Task Definition}
In this section, we first provide the formal task definition for context-dependent text-to-SQL parsing.
Let $U=\{u_1,\ldots, u_T\}$ denote the utterances in a context-dependent text-to-SQL conversation with $T$ turns, where $u_i$ represents the $i$-th NL question. Each NL sentence $u_i$ contains $n_i$ tokens, denoted as $u_i=[w_1, \ldots,w_{n_i}]$. In addition, there is a corresponding database schema $s$, which consists of $N$ tables $\{\mathcal{T}_i\}_{i=1}^{N}$. The number of columns of all tables in the schema is $m$. 
We use $s^{i}$ to denote the name of the $i$-th item in schema $s$.
At current turn $t$, the goal of text-to-SQL parsing is to generate the SQL query $o_t$ given the current utterance $u_t$, historical utterances $\{u_1,\ldots,u_{t-1}\}$, schema $s$, and the last predicted SQL query $o_{t-1}$. 
\name primarily consists of a stack of Transformer layer, which converts a sequence of $L$ input tokens $x=[x_1, ..., x_L]$ into a sequence of contextualized vector representations $\mathbf{h}=[\mathbf{h}_1, \ldots, \mathbf{h}_L]$. 

\section{Pre-training Objectives}

As illustrated in Figure \ref{fig:overview}, we propose two novel pre-training objectives \textbf{SST} (\textbf{S}chema \textbf{S}tate \textbf{T}racking) and \textbf{UDT} (\textbf{U}tterance \textbf{D}ependency \textbf{T}racking) to explore the complex context interactions of NL utterances and SQL queries within each text-to-SQL conversation, respectively. 
In addition, we also employ the \textbf{MLM} (\textbf{M}asked \textbf{L}anguage \textbf{M}odeling) objective to help learn better contextual representations of the conversations. 
Next, we will introduce the pre-training objectives in detail. 

\subsection{Schema State Tracking}
\label{sec:sst}
The usage of context SQL information contributes greatly to model the current SQL query.
Inspired by the dialogue state tracking \cite{ouyang2020dialogue,wang2021tracking} which keeps track of user intentions in the form of a set of dialogue states (\ie, slot-value pairs) in task-oriented dialogue systems, we propose a schema state tracking (SST) objective in a self-supervised manner to keep track of schema states (or user requests) of context-dependent SQL queries, which aims to predict the values of the schema slots.
Concretely, we track the interaction states of the text-to-SQL conversation in the form of schema-states whose slots are column names of all tables in the schema and their values are from SQL keywords. Taking the SQL query in Figure \ref{fig:sql sim} as example, the value of the schema slot \texttt{[cars\_data]} is the SQL keyword \texttt{[SELECT]}. 

Formally, we first convert the last predicted SQL query $o_{t-1}$ into a set of schema states. Since the names of schema states are names of all schema, the values of those schema states that do not appear in the last SQL query $o_{t-1}$ are set to \texttt{[NONE]}, as shown in Figure \ref{fig:sql sim}. We represent the SQL query $o_{t-1}$ with $m$ schema-states $\{(s_{t-1}^i, v_{t-1}^i)\}_{i=1}^{m}$, where $s_{t-1}^{i}$ denotes the schema-state slot, $v_{t-1}^{i}$ denotes the schema-state value of the slot $s_{t-1}^{i}$, and $m$ represents the number of schema. At the $t$-th turn, the goal of SST is to predict the value $v_t^i$ of each schema-state slot $s_t^i$ of the $t$-th SQL query given all the history utterances $\{u_1,\ldots,u_{t-1}\}$, the current utterance $u_t$ and the schema-states $\{(s_{t-1}^i, v_{t-1}^i)\}_{i=1}^{m}$ of the late query $o_{t-1}$. That is, at the $t$-th turn, the input $I_{t}$ of the SST task is as:
\begin{equation}
    \label{eq1}
    I_t = \big[\{u_1,\ldots,u_{t}\};\{(s_{t-1}^i, v_{t-1}^i)\}_{i=1}^{m}\big]
\end{equation}
Note that the SQL queries within a conversation share the same schema $s$, thus the schema-states of the $t$-th and $t-1$-th SQL queries have the same schema-state slots (i.e., $s_{t-1}^i=s_{t}^i=s^i$).

Since each schema state $c_{t-1}^i=(s_{t-1}^i,v_{t-1}^i)$ contains multiple words, we apply an attentive layer to obtain the representation of $c_{t-1}^i=(s_{t-1}^i,v_{t-1}^i)$. Concretely, given the output contextualized representation $\mathbf{h}^{c_{t-1}^i}_{t}=[\mathbf{h}_{t}^l,\ldots,\mathbf{h}_{t}^{l+|c_{t-1}^i|-1}]$ ($l$ is the start index of $c_{t-1}^i$)
of each schema state $c_{t-1}^i$, the attentive schema-state representation $\mathbf{c}_{t-1}^i$ of the schema state $c_{t-1}^i$ can be calculated as:
\begin{gather}
    \alpha_{t-1}^j = \textrm{softmax}\;(\textrm{tanh}(\mathbf{h}_t^{l+j} \mathbf{W}_1)\mathbf{v}_1^{\top})\\
    \mathbf{c}_{t-1}^i = \sum_{j=1}^{|c_{t-1}^i|} \alpha_{t-1}^j \mathbf{h}_t^{l+j}
\end{gather}
where $\mathbf{v}_1$ and $\mathbf{W}_1$ are trainable parameters.
We use the attentive schema-state representation  $\mathbf{c}_{t-1}^i$ in the last SQL query to predict the value $v_t^i$ of the current schema state $\mathbf{c}_t^i$:
\begin{equation}
    P(\mathbf{c}_t^i|\mathbf{c}_{t-1}^i) = \textrm{softmax}(\mathbf{W}_2 \mathbf{c}_{t-1}^i + \mathbf{b}_2)
\end{equation}
where $\mathbf{W}_2$ and $\mathbf{b}_2$ are trainable parameters.

Finally, the pre-training loss function of SST is defined as the cross-entropy between the predicted schema-state value $ P({v}_t^i|\mathbf{c}_{t-1}^i)$ and the gold schema-state value ${v}_t^i$ as follows:
\begin{equation}
    \mathcal{L}_{\rm SST}= -\frac{1}{m} \sum_{i=1}^{m} \mathbf{c}_t^i \log P(\mathbf{c}_t^i|\mathbf{c}_{t-1}^i)
\end{equation}
where $m$ is the number of slot (schema).

\subsection{Utterance Dependency Tracking}
We propose an utterance dependency tracking (UDT) objective to capture complex semantic dependency of sequential NL questions within each text-to-SQL conversation.
A key challenge behind UDT is how to construct appropriate positive and negative labels by way of self-supervision.

Generally, it is intuitive that we can construct negative utterance pairs by selecting NL utterances from different conversations. However, it is non-trivial to construct positive utterance pairs, since the current utterance may be irrelevant to those of the historical utterances with prominent contextual shifts, as the second and third utterances shown in Figure 1.
Hence, we treat the NL utterances within the same conversation as positive pairs, which are assigned with different similarity scores. 
SQL is a highly structured indication of user utterance, so by measuring the similarity of current SQL to historical SQL, pseudo-labels of utterance semantic dependencies can be obtained to guide the \name in contextual modelling.
Here we propose a method to measure SQL similarity from two perspectives.

\begin{figure*}
    \centering
    \includegraphics[width=0.95\textwidth]{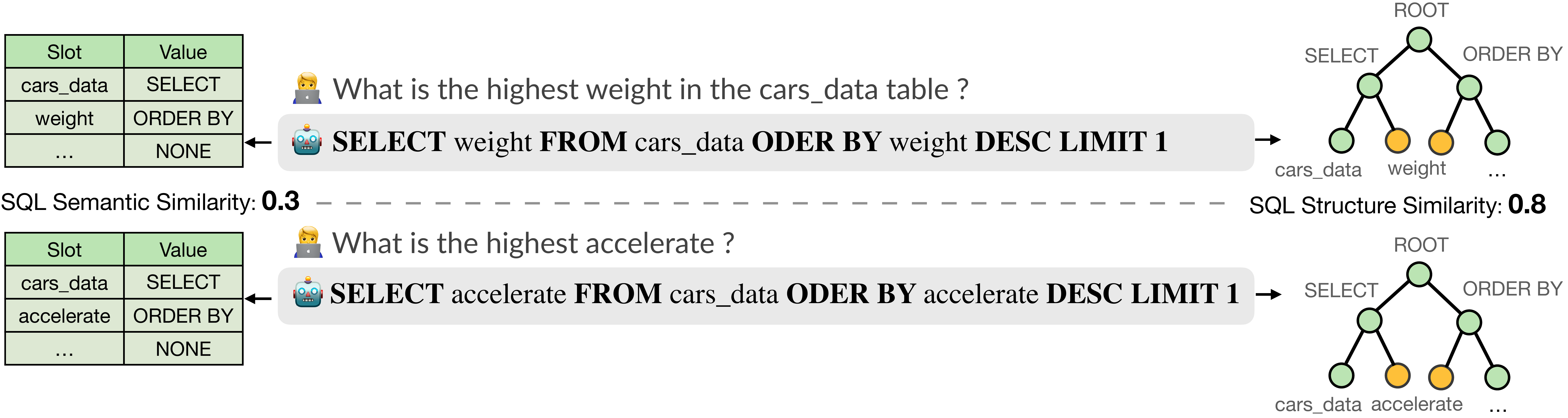}
    \caption{Two metrics for calculating SQL similarity, including semantic similarity and structure similarity.}
    \label{fig:sql sim}
\end{figure*}

\paragraph{SQL Semantic Similarity}
To compute the similarity of two SQL queries, we first convert each SQL query into $m$ schema-states as described in Section \ref{sec:sst}, where the schema slots are names of all schema and their values are from SQL keywords. As illustrated in Figure \ref{fig:sql sim}, given two SQL queries (denotes as $o_x$ and $o_y$), we obtain the schema states  
$\{(s_{x}^i, v_{x}^i)\}_{i=1}^{m}$ and $\{(s_{y}^i, v_{y}^i)\}_{i=1}^{m}$ of the SQL queries $o_x$ and $o_y$ respectively. Since all the schema-states share the same schema slots, we have $s_{x}^i=s_{y}^i$.  
Then, we adopt the Jaccard similarity \cite{niwattanakul2013using} to compute the semantic similarity of the SQL queries $o_x$ and $o_y$ by comparing $v_{x}^i$ and $v_{y}^i$. Mathematically, we compute the SQL semantic similarity of $o_x$ and $o_y$ as:
\begin{equation}
    f_{\rm semantic}(o_x,o_y) = \frac{\sum_{i=1}^{m}{\rm Jaccard}(v_x^i, v_y^i)}{|\hat{s}_{x,y}|}\\
\end{equation}
\begin{equation}
    {\rm Jaccard}(v_x^i,v_y^i) = \frac{|v_x^i\cap v_y^i|}{|v_x^i\cup v_y^i|}
\end{equation}
where $|\hat{s}_{x,y}|$ represents the number of non-duplicate schema states whose values are not \texttt{[NONE]} in $o_x$ and $o_y$. Jaccard function computes the ratio of intersection over the union of $v_x^i$ and $v_y^i$. 

\paragraph{SQL Structure Similarity}
To take advantage of the tree-structure of SQL queries, we first parse each SQL query $o_x$ into a SQL tree $G_x$ as illustrated in Figure \ref{fig:sql sim}. Given two SQL trees $G_x$ and $G_y$ for SQL queries $o_x$ and $o_y$, we leverage the Weisfeiler-Lehman sub-tree kernel \cite{shervashidze2011weisfeiler} to compute the SQL tree-structure similarity score $f_{\rm tree}(o_x,o_y)$ as follows:
\begin{gather}
 f_{\rm tree}(o_x,o_y) = \textrm{Norm}(\mathcal{K}_{\rm WL}(\mathbf{G}_{x}, \mathbf{G}_{y})) \\ 
    \mathcal{K}_{\rm WL}(\mathbf{G}_{x},\mathbf{G}_{y}) = \sum_{i=0}^{h} \mathcal{K}(\mathbf{G}^{i}_{x},\mathbf{G}^{i}_{y}) 
\end{gather}
where $\textrm{Norm}$() is a normalization function, $\mathcal{K}_{\rm WL}()$ is the Weisfeiler-Lehman subtree kernel function and $\mathcal{K}()$ is the base kernel on graphs. $\mathbf{G}^{i}_{x}$ denotes the Weisfeiler-Lehman graph at height $i$ of the tree $\mathbf{G}_{x}$ and $h$ is the number of Weisfeiler-Lehman iterations. We refer the readers to \citet{shervashidze2011weisfeiler} for the implementation details of Weisfeiler-Lehman sub-tree kernel.

Overall, we define the final similarity score of two SQL queries $o_x$ and $o_y$ as follows:
\begin{equation}
\begin{aligned}
     f_{\rm SQL}(o_x,o_y) = & \lambda \cdot  f_{\rm semantic}(o_x,o_y) \\
    &+ (1-\lambda) \cdot f_{\rm tree}(o_x,o_y)
    \end{aligned}
\end{equation}
where $\lambda$ is a hyper-parameter controlling the impact of the two kinds of similarity.

\paragraph{Weighted Contrastive Loss}
After obtaining the SQL similarity, we employ weighted contrastive learning ~\cite{oord2018representation} to pull together two semantically similar NL utterances and push away the representations of semantically dissimilar NL utterances within each conversation. 
We first convert the input sequence $I_t=[x_t^1,\ldots,x_t^L]$ into a sequence of contextualized vectors $\mathbf{h}_t=[\mathbf{h}_t^1,\ldots,\mathbf{h}_t^L]$, where $L$ represents the length of the input sequence. We leverage an attention mechanism to learn the input representation $\mathbf{\tilde{h}}_t$ as:
\begin{gather}
    \beta_t^i = {\rm Softmax}\;({\rm tanh}(\mathbf{h}_t^i \mathbf{W}_3)\mathbf{v}_3^{\top})\\
    \mathbf{\widetilde{h}}_t = \sum_{i=1}^{L} \beta_t^i \mathbf{h}_t^{i}
\end{gather}
where $\mathbf{v}_3$ and $\mathbf{W}_3$ are trainable parameters. 

Specifically, we minimize a weighted contrastive loss function $\mathcal{L}_{\rm UDT}$ to optimize the  network as:
\begin{equation}
\small
\begin{aligned}
    \mathcal{L}_{\rm UDT} & = -\sum_{x\in \mathcal{D}}\sum_{p\in \mathcal{D}_{x}^{+}} \frac{f_{\rm SQL}(o_x, o_p)}{\sum_{k\in \mathcal{D}}f_{\rm SQL}(o_x, o_k)}\\
    & \cdot {\rm log}\frac{e^{\text{sim}(\mathbf{\tilde{h}}_x,\mathbf{\tilde{h}}_p)/\tau}}{e^{\text{sim}(\mathbf{\tilde{h}}_x,\mathbf{\tilde{h}}_p)/\tau} + \sum_{m\in \mathcal{D}^{-}_{x}}e^{\text{sim}(\mathbf{\tilde{h}}_x,\mathbf{\tilde{h}}_m)/\tau}}
\end{aligned}
\end{equation}
where $\tau$ is a temperature hyper-parameter. 
$D=\{1,\ldots,N\}$ denotes the index set of the training utterances. $\mathcal{D}^{+}_{x}$ denotes the index set of positive utterances that co-occurs in the same conversation with utterance $x$.  $\mathcal{D}^{-}_{x}$ denotes the index set of positive utterances other than $x$ and $p$, and negative utterances chosen from other conversations. 

\subsection{Masked Language Modeling}
In order to jointly learn the contextual representation of utterances and schema, we retain the masking mechanism in the pre-training stage. 
Concretely, given the input $I_t$ $\label{eq:input}$ (defined in Eq.\ref{eq1}) of the $t$-th turn, masked language modeling (MLM) selects a random set of positions and replaces these positions with \texttt{[MASK]}, and then learns to predict the original tokens of the masked-out tokens. 
We follow the hyperparameters of prior work \cite{devlin2018bert}, which randomly masks utterances and schema tokens with a 15\% probability.
We denote the MLM loss as $\mathcal{L}_{\rm MLM}$, which is computed by minimizing the cross-entropy function on the masked tokens.
\subsection{Joint Pre-training Objective}
In this paper, we combine three pre-training objectives to learn a pre-training framework for context-dependent text-to-SQL parsing. Instead of combining the objectives by simply performing a weighted linear sum of individual losses, we jointly learn three objectives by considering the homoscedastic uncertainty of each objective \cite{Kendall_2018_CVPR}. In this way, we can avoid the huge expense to tune weight hyper-parameters. We define the joint loss function based on homoscedastic
uncertainty as:
\begin{equation}
\small
\begin{aligned}
    & \mathcal{L}_{\rm joint} = \frac{1}{2\sigma_1^2}\mathcal{L}_{\rm SST} + \frac{1}{2\sigma_2^2}\mathcal{L}_{\rm UDT} + \frac{1}{2\sigma_3^2}\mathcal{L}_{\rm MLM}\\
    &~~ + \log(1+\sigma_1) + \log(1+\sigma_2) + \log(1+\sigma_3) 
\end{aligned}
\end{equation}
where $\sigma_1, \sigma_2, \sigma_3$ represent the model's observation noise parameters, capturing how much noise we have in the outputs.

\section{Data Construction for Pre-training}
The cost of expensive SQL annotation poses a challenge to the construction of large scale pre-training data.
Previous work \citep{yu2021grappa,yu2020score} resort to data augmentation to address this issue.
Typically in a conversational setting, context-dependent data augmentation techniques require two steps: (1) single-turn context-free grammar for utterance-SQL pair generation, and (2) a follow-up context-free grammar to expand single-turn data into context-dependent conversations. 
\score synthesized a total of 435k text-to-SQL conversations following this setup, and we noticed two limitations with it. 
Firstly, it relies on the template-filling construction to convert SQL to utterances, resulting in rather rigid generated utterances in step (1). 
Secondly, \sparc is the only data resource employed to induce the follow-up context-free grammar in step (2). 
Nevertheless, the contextual diversity in \sparc is insufficient to simulate complex contextual dependencies.

To this end, we propose a new pre-training data construction method.
Inspired by the SNOWBALL framework \cite{shu2021logic}, we harness a generative model, i.e., BART, to bring more diversity to the generated utterances.
For the follow-up conversational context-free grammar induction, we consider both \cosql and \sparc datasets and manually craft 100 templates.
Overall, we synthesize a new large-scale pre-training dataset that consists of about 480K high-quality context-dependent text-to-SQL conversations.
We provide examples of the induced grammar rules and synthesized procedure in detail in Appendix \ref{sec:data-constraction}.

\begin{table*}[!htbp]
    \small
	\centering
	\scalebox{1.0}{
	\begin{tabular}{l|cccc|cccc}
		\toprule 
		{\multirow{3}*{\textbf{Model}}} & \multicolumn{4}{c|}{\textbf{\sparc}} & \multicolumn{4}{c}{\textbf{\cosql}} \\
		\cmidrule{2-9}
		\multirow{2}{*}{} &
		\multicolumn{2}{c}{QM} &
		\multicolumn{2}{c|}{IM} &
		\multicolumn{2}{c}{QM} &
		\multicolumn{2}{c}{IM} \\
		& Dev & Test & Dev & Test & Dev & Test & Dev & Test  \\
		\midrule
		\multicolumn{9}{c}{\textit{Previous Parsing Systems.}} \\
		\midrule
		GAZP + \bert & 48.9 & 45.9 & 29.7 & 23.5 & 42.0 & 39.7 & 12.3 & 12.8\\
		EditSQL + \bert & 47.2 & 47.9 & 29.5 & 25.3 & 39.9 & 40.8 & 12.3 & 13.7  \\
		IGSQL + \bert & 50.7 &  51.2 &  32.5 & 29.5 & 44.1 & 42.5 & 15.8 & 15.0\\
		IST-SQL + \bert & 47.6 & - & 29.9 & - & 44.4 & 41.8 & 14.7 & 15.2\\
		$\text{R}^{2}$SQL + \bert & 54.1 &  55.8 & 35.2 &  30.8 & 45.7 & 46.8 & 19.5 & 17.0 \\
		DELTA + \bart & 58.6 & 59.9 & 35.6 & 31.8 & 51.7 & 50.8 & 21.5 & 19.7 \\
		RAT-SQL + \score & 62.2 &  62.4 &  42.5 & 38.1 & 52.1 & 51.6 & 22.0 & 21.2 \\
		T5-3B + \picard & - & - & - & - & 56.9 & 54.6 & 24.2 & 23.7 \\
		HIE-SQL + $\grappa $ & 64.7 & 64.6 & 45.0 & 42.9 & 56.4 & 53.9 & 28.7 & 24.6 \\
		\midrule
		\multicolumn{9}{c}{\textit{Pre-trained Models.}} \\
		\midrule
		LGESQL & 52.4 & - & 31.3 & - & 41.2 & - & 15.0  & -\\
		\quad \textit{w.} \bert & 59.8 & - & 40.5 & - & 50.7 & - & 20.8  & -\\
		\quad \textit{w.} \roberta & 61.6 & - & 41.2 & - & 51.9 & - & 20.8  & -\\
		\quad \textit{w.} \grappa & 62.5 & - & 42.4 & - & 52.6 & - & 21.5  & -\\
		\quad \textit{w.} \score & 62.3 & - & 43.6 & - & 52.3 & - & 22.5  & -\\
		\rowcolor[RGB]{237,237,237} \quad \textit{w.} \textbf{\name} & \textbf{66.9} & \textbf{67.4} ($\uparrow$ 2.8) & \textbf{46.9} & \textbf{46.6} ($\uparrow$ 3.7) & \textbf{59.7} & \textbf{57.8} ($\uparrow$ 3.9) & \textbf{30.0}  & \textbf{28.2} ($\uparrow$ 3.6)\\
		\bottomrule
	\end{tabular}
	}
	\caption{Experimental results of various methods in terms of question match (QM) accuracy and interaction match (IM) accuracy on both  \sparc and \cosql datasets. ``-” means that the test results are not accessible since the test accuracy needs to be officially evaluated and only two models can be submitted every two months.}
    \label{tab:main-result}
\end{table*}

\begin{table}[t]
    \small
	\centering
	\scalebox{0.9}{
	\begin{tabular}{l|cc|cc}
		\toprule 
		{\multirow{3}*{\textbf{Model}}} & \multicolumn{2}{c|}{\textbf{\sparc}} & \multicolumn{2}{c}{\textbf{\cosql}} \\
		\cmidrule{2-5}
		\multirow{2}{*}{} &
		\multicolumn{1}{c}{QM} &
		\multicolumn{1}{c|}{IM} &
		\multicolumn{1}{c}{QM} &
		\multicolumn{1}{c}{IM} \\
		\midrule
		\name & \textbf{66.9} & \textbf{46.9} & \textbf{59.7} & \textbf{30.0} \\
		\quad \textit{w/o} MLM & 66.1 & 45.7 & 59.0 & 28.7 \\
		\quad \textit{w/o} SST & 66.8 & 45.5 & 57.9 & 28.3 \\
		\quad \textit{w/o} UDT & 66.4 & 46.1 & 58.0 & 28.7 \\
		\quad \textit{w/o} SST+UDT & 65.3 & 45.6 & 57.0 & 27.3 \\
		\bottomrule
	\end{tabular}
	}
	\caption{Ablation study of \name in terms of question match accuracy (QM) and interaction match accuracy (IM) on the dev sets of both \sparc and \cosql.}
	\label{ablation}
\end{table}

\section{Experiment}
\subsection{Experimental Setup}
\paragraph{Downstream Datasets} 
We evaluate \name on two context-dependent semantic parsing benchmarks: \sparc \cite{yu2019sparc} and \cosql \cite{yu2019cosql}. \sparc is a collection of cross-domain context-dependent dataset, which consists of about 4.3k  question sequences and 12k+ individual questions annotated with SQL queries. \cosql is a conversational text-to-SQL corpus, which contains about 3k dialogues and 10k+ annotated SQL queries.  Both \sparc and \cosql query 200
complex databases spanning across 138 domains.
We provide more detailed statistics of these two datasets in Appendix \ref{sec:data-details}.

\paragraph{Evaluation Metrics} 
We employ two official evaluation metrics \cite{yu2019sparc,yu2019cosql} to verify the effectiveness of \name: question match accuracy (QM) and interaction match accuracy (IM). Concretely, QM denotes the exact set match accuracy over SQL templates and IM denotes the ratio of interactions over all correctly predicted questions.

\paragraph{Implementation Details}
In pre-training, \name is initialized with \electra \cite{clark2020electra}. Similar to ELECTRA, we also employ the replaced token detection objective to further improve the text-to-SQL pre-training. The maximum length of each input sequence is set to 256. The batch size is set to 80 and an Adam optimizer is employed for optimization with an initial learning rate of 1e-6. Gradient clipping is applied to \name with a maximum gradient value of 1. For computing the SQL similarity, the impact factor $\lambda$ is set to 0.5. 
We provide more details of implementation in Appendix \ref{sec:more-implementation-details}.

\paragraph{Baselines}
First, we compare \name with several state-of-the-art context-dependent parsing methods, including \textbf{GAZP} \cite{zhong2020grounded}, \textbf{EditSQL} \cite{zhang2019editing}, \textbf{IGSQL} \cite{cai2020igsql}, \textbf{IST-SQL} \cite{wang2021tracking}, \textbf{$\text{R}^{2}$SQL} \cite{hui2021dynamic}, \textbf{\picard} \cite{scholak2021picard}, 
\textbf{DELTA} \cite{chen2021decoupled}
and \textbf{HIE-SQL} \cite{zheng2022hie}. 
Second, we compare \name with four strong pre-training models, including  \textbf{\bert} \cite{devlin2018bert}, \textbf{\textsc{RoBERTa}} \cite{liu2019roberta}, \textbf{\grappa} \cite{yu2021grappa} and \textbf{\score} \cite{yu2020score}. In particular, \grappa and \score are the representative TaLMs for context-independent and context-dependent text-to-SQL parsing, respectively.

\pgfplotstableread[row sep=\\,col sep=&]{
    Scale & w/o PLM & score & grappa & name \\
    easy  & 67.3 & 75.8 & 78.5 & 78.7 \\
    medium  & 51.2 & 62.1 & 59.6 & 68.9\\
    hard   & 33.8 & 35.9 & 41.4 & 45.5 \\
    extra hard & 22.4  & 30.6 & 32.1 & 41.0 \\
    }\mydata

\pgfplotsset{every axis/.append style={
                    label style={font=\large},
                    tick label style={font=\large} 
                    }}
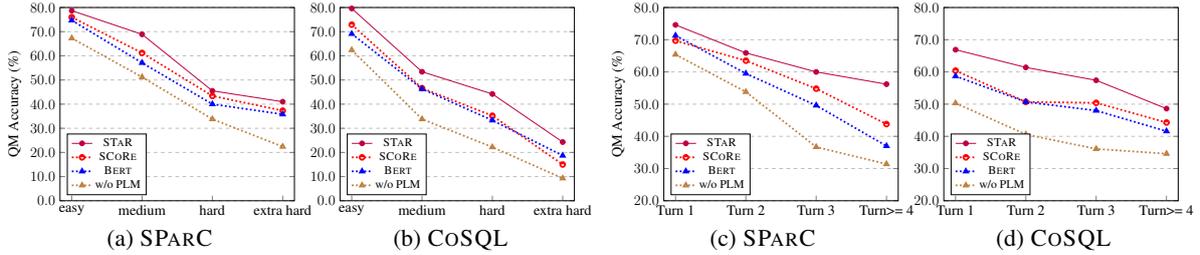
\begin{figure*}[t]
    \begin{minipage}{0.23\linewidth}
    \centering
            \begin{tikzpicture}[scale=0.45]
                \begin{axis}[
                    legend style={nodes={scale=0.9, transform shape}},
                    xtick pos=left,
                    log ticks with fixed point,
                    ylabel={QM Accuracy (\%)},
                    xmin=0.00, xmax=1.0,
                    ymin=0.0, ymax=0.8,
                    xtick={0.05, 0.35, 0.65, 0.95},
                    xticklabels={easy, medium, hard, extra hard},
                    ytick={0.0, 0.10, 0.20, 0.30, 0.4, 0.5, 0.6, 0.7, 0.8},
                    yticklabels={$0.0$, $10.0$, $20.0$, $30.0$,  $40.0$, $50.0$, $60.0$, $70.0$, $80.0$},
                    legend pos=south west,
                    ymajorgrids=true,
                    grid style=dashed
                ]
             
                \addplot[
                    color=purple,
                    mark options={solid},
                    mark=*,
                    mark size=2pt
                    ]
                    coordinates {
                    (0.05, 0.787)
                    (0.35, 0.689)
                    (0.65, 0.455)
                    (0.95, 0.410)
                    };            
                    \addlegendentry{\name}
                
                \addplot[
                    color=red,
                    dotted,
                    mark=halfcircle*,
                    mark options={solid},
                    line width=1.5pt,
                    mark size=2pt
                    ]
                    coordinates {
                    (0.05, 0.760)
                    (0.35, 0.612)
                    (0.65, 0.434)
                    (0.95, 0.373)
                    };            
                    \addlegendentry{\score}

                \addplot[
                    color=blue,
                    dotted,
                    mark=triangle*,
                    mark options={solid},
                    line width=1.5pt,
                    mark size=2pt
                    ]
                    coordinates {
                    (0.05, 0.747)
                    (0.35, 0.571)
                    (0.65, 0.400)
                    (0.95, 0.358)
                    };            
                    \addlegendentry{\bert}    
                 
                \addplot[
                    color=brown,
                    dotted,
                    mark=triangle,
                    mark options={solid},
                    line width=1.5pt,
                    mark size=2pt
                    ]
                  coordinates {
                    (0.05, 0.673)
                    (0.35, 0.512)
                    (0.65, 0.338)
                    (0.95, 0.224)
                    };            
                    \addlegendentry{w/o PLM}
                \end{axis}
                \node[below,font=\small] at (current bounding box.south) {(a) \sparc};
                \end{tikzpicture}
    \end{minipage}
    \hspace{0.1cm}
    \begin{minipage}{0.23\linewidth}
    \centering
            \begin{tikzpicture}[scale=0.45]
                \begin{axis}[
                    legend style={nodes={scale=0.9, transform shape}},
                    xtick pos=left,
                    log ticks with fixed point,
                    xmin=0.00, xmax=1.0,
                    ymin=0.00, ymax=0.8,
                    xtick={0.05, 0.35, 0.65, 0.95},
                    xticklabels={easy, medium, hard, extra hard},
                    ytick={0.0, 0.10, 0.20, 0.30, 0.4, 0.5, 0.6, 0.7, 0.8},
                    yticklabels={$0.0$, $10.0$, $20.0$, $30.0$,  $40.0$, $50.0$, $60.0$, $70.0$, $80.0$},
                    legend pos=south west,
                    ymajorgrids=true,
                    grid style=dashed
                ]
            
                \addplot[
                    color=purple,
                    mark options={solid},
                    mark=*,
                    mark size=2pt
                    ]
                    coordinates {
                    (0.05, 0.796)
                    (0.35, 0.534)
                    (0.65, 0.442)
                    (0.95, 0.243)
                    };            
                    \addlegendentry{\name}

                \addplot[
                    color=red,
                    dotted,
                    mark=halfcircle*,
                    mark options={solid},
                    line width=1.5pt,
                    mark size=2pt
                    ]
                    coordinates {
                    (0.05, 0.729)
                    (0.35, 0.466)
                    (0.65, 0.352)
                    (0.95, 0.150)
                    };            
                    \addlegendentry{\score}
                     
                \addplot[
                    color=blue,
                    dotted,
                    mark=triangle*,
                    mark options={solid},
                    line width=1.5pt,
                    mark size=2pt
                    ]
                    coordinates {
                    (0.05, 0.691)
                    (0.35, 0.463)
                    (0.65, 0.333)
                    (0.95, 0.187)
                    };            
                    \addlegendentry{\bert}    
                 
                \addplot[
                    color=brown,
                    dotted,
                    mark=triangle,
                    mark options={solid},
                    line width=1.5pt,
                    mark size=2pt
                    ]
                  coordinates {
                    (0.05, 0.624)
                    (0.35, 0.338)
                    (0.65, 0.222)
                    (0.95, 0.093)
                    };            
                    \addlegendentry{w/o PLM}
                \end{axis}
                \node[below,font=\small] at (current bounding box.south) {(b) \cosql};
                \end{tikzpicture}

    \end{minipage}
    \hspace{0.1cm}
    \begin{minipage}{0.23\linewidth}
    
    \centering
            \begin{tikzpicture}[scale=0.45]
                \begin{axis}[
                    legend style={nodes={scale=0.9, transform shape}},
                    xtick pos=left,
                    log ticks with fixed point,
                    ylabel={QM Accuracy (\%)},
                    xmin=0.00, xmax=1.0,
                    ymin=0.2, ymax=0.8,
                    xtick={0.05, 0.35, 0.65, 0.95},
                    xticklabels={Turn 1, Turn 2, Turn 3, Turn>= 4},
                    ytick={0.20, 0.30, 0.4, 0.5, 0.6, 0.7, 0.8},
                    yticklabels={$20.0$, $30.0$,  $40.0$, $50.0$, $60.0$, $70.0$, $80.0$},
                    legend pos=south west,
                    ymajorgrids=true,
                    grid style=dashed
                ]
                 
                \addplot[
                    color=purple,
                    mark options={solid},
                    mark=*,
                    mark size=2pt
                    ]
                    coordinates {
                    (0.05, 0.746)
                    (0.35, 0.659)
                    (0.65, 0.600)
                    (0.95, 0.562)
                    };            
                    \addlegendentry{\name}

                \addplot[
                    color=red,
                    dotted,
                    mark=halfcircle*,
                    mark options={solid},
                    line width=1.5pt,
                    mark size=2pt
                    ]
                    coordinates {
                    (0.05, 0.697)
                    (0.35, 0.635)
                    (0.65, 0.548)
                    (0.95, 0.438)
                    };            
                    \addlegendentry{\score}
                    
                \addplot[
                    color=blue,
                    dotted,
                    mark=triangle*,
                    mark options={solid},
                    line width=1.5pt,
                    mark size=2pt
                    ]
                    coordinates {
                    (0.05, 0.713)
                    (0.35, 0.595)
                    (0.65, 0.496)
                    (0.95, 0.370)
                    };            
                    \addlegendentry{\bert}    
                
                \addplot[
                    color=brown,
                    dotted,
                    mark=triangle,
                    mark options={solid},
                    line width=1.5pt,
                    mark size=2pt
                    ]
                  coordinates {
                    (0.05, 0.654)
                    (0.35, 0.538)
                    (0.65, 0.367)
                    (0.95, 0.314)
                    };            
                    \addlegendentry{w/o PLM}
                \end{axis}
                \node[below,font=\small] at (current bounding box.south) {(c) \sparc};
                \end{tikzpicture}
            
            \label{levels}
    \end{minipage}
    \hspace{0.1cm}
    \begin{minipage}{0.23\linewidth}
    \centering
            \begin{tikzpicture}[scale=0.45]
                \begin{axis}[
                    legend style={nodes={scale=0.9, transform shape}},
                    xtick pos=left,
                    log ticks with fixed point,
                    xmin=0.0, xmax=1.0,
                    ymin=0.2, ymax=0.80,
                    xtick={0.05, 0.35, 0.65, 0.95},
                    xticklabels={Turn 1, Turn 2, Turn 3, Turn>= 4},
                    ytick={0.20, 0.30, 0.4, 0.5, 0.6, 0.7, 0.8},
                    yticklabels={$20.0$, $30.0$,  $40.0$, $50.0$, $60.0$, $70.0$, $80.0$},
                    legend pos=south west,
                    ymajorgrids=true,
                    grid style=dashed
                ]

                \addplot[
                    color=purple,
                    mark options={solid},
                    mark=*,
                    mark size=2pt
                    ]
                    coordinates {
                    (0.05, 0.669)
                    (0.35, 0.614)
                    (0.65, 0.574)
                    (0.95, 0.486)
                    };            
                    \addlegendentry{\name}

                \addplot[
                    color=red,
                    dotted,
                    mark=halfcircle*,
                    mark options={solid},
                    line width=1.5pt,
                    mark size=2pt
                    ]
                    coordinates {
                    (0.05, 0.604)
                    (0.35, 0.507)
                    (0.65, 0.504)
                    (0.95, 0.443)
                    };            
                    \addlegendentry{\score}
                    
                \addplot[
                    color=blue,
                    dotted,
                    mark=triangle*,
                    mark options={solid},
                    line width=1.5pt,
                    mark size=2pt
                    ]
                    coordinates {
                    (0.05, 0.587)
                    (0.35, 0.507)
                    (0.65, 0.480)
                    (0.95, 0.416)
                    };            
                    \addlegendentry{\bert}    

                \addplot[
                    color=brown,
                    dotted,
                    mark=triangle,
                    mark options={solid},
                    line width=1.5pt,
                    mark size=2pt
                    ]
                  coordinates {
                    (0.05, 0.503)
                    (0.35, 0.406)
                    (0.65, 0.361)
                    (0.95, 0.346)
                    };            
                    \addlegendentry{w/o PLM}
                \end{axis}
                \node[below,font=\small] at (current bounding box.south) {(d) \cosql};
                \end{tikzpicture}
        
            \label{levels}
    \end{minipage}
    \caption{The results of \name and baselines on \sparc and \cosql dev sets (a-b) by varying the difficulty levels of the data and (c-d) by varying the conversation turns.} 
    \label{fig:diff}
\end{figure*}

\subsection{Model Comparison on Downstream Tasks}
In the experiments, we choose LGESQL \cite{cao2021lgesql} as our base model given its superior performance. Since LGESQL is originally developed for single-turn setting, we extend LGESQL to context-dependent setting by taking as input the concatenation of historical and current utterances. For a fair comparison, the four compared PLMs also leverage LGESQL as the base model.  

The experimental results on \sparc and \cosql are summarized in Table \ref{tab:main-result}.  \name outperforms all the compared methods on the two datasets by a noticeable margin. First, \name achieves substantially better results than the four strong PLMs. In particular, \name surpasses the well-known \score by 7.4\% QM score and 7.5\% IM score on the \cosql dev set.
Second, LGESQL+\name achieves better results than the compared downstream methods which use \bert, \roberta, \score, \grappa as the PLMs, such as the best performing baseline HIE-SQL+\grappa. 
\begin{table}[t]
    \small
	\centering
	\scalebox{0.85}{
	\begin{tabular}{l|cc|cc}
		\toprule 
		{\multirow{3}*{\textbf{Model}}} & \multicolumn{2}{c|}{\textbf{\cosql}} & \multicolumn{2}{c}{\textbf{\sparc}} \\
		\cmidrule{2-5}
		\multirow{2}{*}{} &
		\multicolumn{1}{c}{QM} &
		\multicolumn{1}{c|}{IM} &
		\multicolumn{1}{c}{QM} &
		\multicolumn{1}{c}{IM} \\
		\midrule
		\name  & 59.7 & 30.0 & 66.9 & 46.9 \\
		\name w/o structural & 59.1 & 29.0 & 66.5 & 46.7 \\
		\name w/o semantic & 59.5 & 29.6 & 66.8 & 46.5 \\
		\name w/o UDT & 58.0 & 28.6 & 66.4 & 46.1 \\
		\bottomrule
	\end{tabular}
	}
	\caption{Results of \name  on the dev sets of \sparc and \cosql by using different metrics for calculating SQL similarity.}
	\label{metric}
\end{table}
\subsection{Ablation Study}

\paragraph{Effectiveness of Pre-training Objectives} We conduct ablation test to investigate the effectiveness of each pre-training objective in \name. We report the results of removing the MLM loss (called w/o MLM), the SST loss (called w/o SST), the UDT loss (called w/o UDT), and both SST and UDT (called w/o SST+UDT) respectively. Table \ref{ablation} shows the ablation test results on both \sparc and \cosql. We can observe that removing the SST or UDT objective bring the most significant performance drop. Not surprisingly, combining all the three objectives  achieves the best results on both datasets.

\paragraph{Effectiveness of SQL Similarity Metrics}
To analyze the impact of metrics for calculating the SQL similarity in \name, we also conduct an ablation test by removing the structural similarity metric (called w/o structural), the semantic similarity metric (called w/o semantic), and both (called w/o UDT), respectively. Table \ref{metric} shows the ablation test results on the dev sets of \sparc and \cosql. As expected, both similarity metrics contribute great improvements to \name.

\paragraph{Effectiveness of Synthesized Pre-training Data}
We also analyze the quality of our constructed pre-training data. We compare our pre-training data with the data created by \score \cite{yu2020score} which to our knowledge is the only existing work on pre-training for context-dependent text-to-SQL parsing. Since the pre-training data created by \score is inapplicable to the $\mathcal{L}_{\rm UDT}$ objective, we merely employ $\mathcal{L}_{\rm MLM}$ (denoted as \name w/ MLM) and  $\mathcal{L}_{\rm MLM}$ + $\mathcal{L}_{\rm SST}$ (denoted as \name w/ MLM + SST) as the pre-training objectives in the experiments. As shown in Table \ref{pretrain-data}, our pre-training data is more effective than the pre-training data created by \score.

\begin{figure*}
    \centering
    \includegraphics[width=1\textwidth]{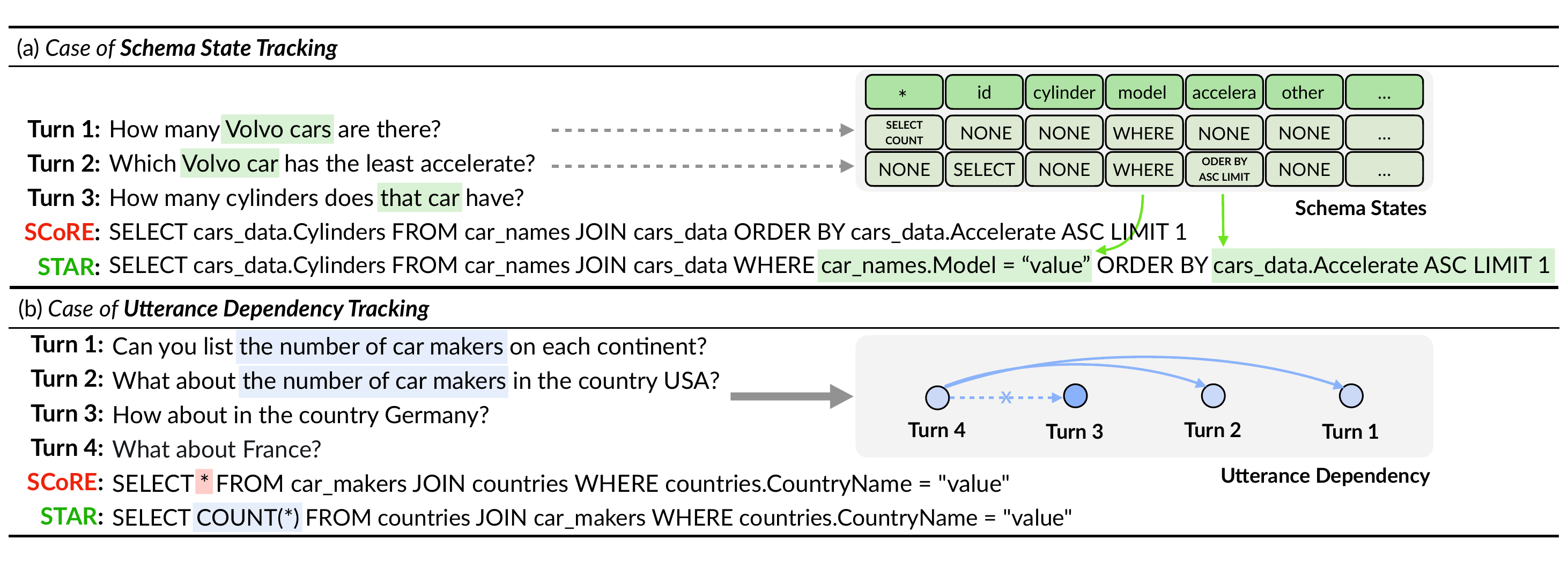}
    \caption{Two cases on the \cosql dev dataset.}
    \label{fig:case study}
\end{figure*}

\subsection{Discussion}

\begin{table}[t]
    \small
	\centering
	\scalebox{0.85}{
	\begin{tabular}{l|cc}
		\toprule 
		{\multirow{3}*{\textbf{Model}}} &  \multicolumn{2}{c}{\textbf{\cosql}} \\
		\cmidrule{2-3}
		\multirow{2}{*}{} &
		\multicolumn{1}{c}{QM} &
		\multicolumn{1}{c}{IM}  \\
		\midrule
		\name (w/ MLM) + \score data & 55.4 & 25.6 \\
		\name (w/ MLM) + Our data & 57.0 & 27.3\\
		\name (w/ MLM + SST) + \score data & 57.3 & 27.3 \\
		\name (w/ MLM + SST) + Our data & 58.0 & 28.7\\
		\bottomrule
	\end{tabular}
	}
	\caption{Results of \name on the dev set of \cosql with MLM and SST objectives by using different pre-training data.}
	\label{pretrain-data}
\end{table}

\paragraph{Model Comparison on Samples with Different Levels of Difficulty}
The SQL queries in both \sparc and \cosql can be further divided into four levels based on the difficulty of the SQL queries: easy, medium, hard, extra hard, which can be used to better evaluate the model performance on different queries. 
As shown in Figure \ref{fig:diff}a-b, \name achieves better results than the compared methods on the four kinds of data, even on the extra hard samples. 

\paragraph{Model Comparison on Samples at Different Turns}
Figure \ref{fig:diff}c-d illustrate the QM results of \name and compared methods along with the increase of conversation turns on \sparc and \cosql dev sets.  The QM results of baselines decrease sharply as the conversation turns increase, while \name achieves much more stable performance even for the third and fourth turns. This suggests that \name can better track and explore the interaction states in history utterances to assist the models to better parse current utterance. 

\subsection{Case Study}
To evaluate \name qualitatively, we choose two exemplary conversations from the CoSQL dev set and illustrate the generated SQL queries by \score and \name in Figure \ref{fig:case study}. In the first case, we observe that \name can exploit the usage of table information in history queries (e.g., \texttt{[car\_names.Model]} to correctly generate the third SQL query, while \score fails to track this kind of schema state. In the second case, \name successfully tracks the long-term utterance dependency between the first and fourth utterances, and generates the correct SQL keyword \texttt{[SELECT COUNT(*)]} in the fourth SQL query by tracking and referring to the query ``the number of'' in the second utterance. However, \score fails to track such long-term dependency with being disturbed by the third utterance. 

\subsection{Limitation Analysis}
To better analyze the limitations of \name, we carry out an analysis of the errors made by \name on the CoSQL dev dataset. We reveal several reasons of the errors, which can be divided into following categories. \textbf{First}, \name fails to select the correct names from table schemas in some hard or extra hard samples, where NL questions use synonyms to refer to tables or columns in SQL queries without the explicit correspondence between NL questions and table schemas. 
One possible solution is to exploit the rich semantic information contained in PLMs to capture the implicit schema linking information via knowledge probing techniques.
\textbf{Second}, for some samples, \name incorrectly inherits part of the previous turn SQL query. 
One possible solution is to design an additional classifier to predict the changes (e.g. \textsc{retain}, \textsc{modify}, \textsc{delete}) between the schema state of the current turn and that of the previous turn.
\textbf{Third}, there are some SQL grammar errors such as the redundancy of \texttt{[WHERE]} clause, repetition of table names, structure error of \texttt{[SELECT NEST]}. The reason may be that the schema state tracking objective only tracks the state of the database schema in conversation, which do not consider the overall grammatical structure of SQL queries. One possible idea is to add an extra objective to predict the general structure of SQL (e.g., abstract syntax tree) so as to capture the overall grammatical structure information of SQL.

\section{Related Work}
\paragraph{Context-dependent Text-to-SQL Parsing}
Most of previous text-to-SQL works focused on the context-independent setting.
Notably, the graph-based parser, \eg, RAT-SQL \cite{wang2020rat}, LGESQL \cite{cao2021lgesql} , S$^2$SQL \cite{Hui2022S2SQLIS}, and the T5-based parser, \eg, \picard \cite{scholak2021picard}, achieving the impressive performance on \spider \cite{yu2018spider}.
In recent years, context-dependent (multi-turn) text-to-SQL parsing has attracted increasing attention due to its broad applications and realistic setting.  \sparc \cite{yu2019sparc} and \cosql \cite{yu2019cosql} are two benchmark datasets for context-dependent text-to-SQL parsing. 
Subsequently, several works~\cite{zhang2019editing,cai2020igsql, wang2021tracking,Wang2021AME,hui2021dynamic,chen2021decoupled,zheng2022hie} were proposed, which consider contextual information or conversation history so as to synthesise the correct SQL query. 
In particular, \citet{zhang2019editing} exploited the conversation history by editing the previous predicted SQL to improve the generation quality.
The schema interaction graph in IGSQL \cite{cai2020igsql} and two kinds of interaction states in IST-SQL \cite{wang2021tracking} are designed to capture the historical schema evolution in context.
Furthermore, \citet{zheng2022hie} improve contextual accuracy by incorporating additional SQL encoders to integrate historical SQL into the input.
In contrast to above works, \name focus on the pre-training stage, expecting to extract general knowledge from large-scale unsupervised or self-supervised data that will be useful for downstream parsing tasks.

\paragraph{Pre-training Models for Text-to-SQL Parsing}
In parallel, tabular language models (TaLMs) have been proposed to simultaneously encode tables and texts, which further improved the results of downstream text-to-SQL parsing tasks. 
For example, \textsc{TaBERT} \cite{yin2020tabert} and \textsc{TaPas}~\cite{herzig-etal-2020-tapas} jointly encoded texts and tables with self-supervised or weakly-supervised objectives, which was trained on a large corpus of tables.
\textsc{StruG} \cite{Deng_2021} proposed a structured-grounded pre-training technique and \textsc{GAP} \cite{Shi2020LearningCR} introduced a generation-augmented pre-training framework to capture the alignment relationship of utterance and table.
Similarly, \grappa \cite{yu2021grappa} introduced a grammar-augmented pre-training framework for text-to-SQL parsing, which explored the schema linking by encouraging the model to identify table schema components that could be grounded to logical form constituents.
\score \cite{yu2020score} was the state-of-the-art pre-training approach for context-dependent text-to-SQL parsing designed to induce representations that captured the switch between the adjacency turns.
Unlike these TaLMs, \name is the first to leverage both historical SQL and complex utterance dependency in the pre-training stage.

\section{Conclusion}
In this paper, we proposed \name, a pre-trained TaLM, which could jointly learn user utterance and table schema representations for context-dependent text-to-SQL conversations. 
\name contained two novel pre-training objectives (schema state tracking and utterance dependency tracking) to explore the complex context interactions of NL utterances and SQL queries within each text-to-SQL conversation, respectively.
We constructed a diverse large-scale context-dependent text-to-SQL conversation corpus to pre-train \name. Experiments demonstrated that \name achieves new state-of-the-art performance on  \sparc and \cosql. 

\section*{Acknowledgements}
My acknowledges: This work was partially supported by National Natural Science Foundation of China (No. 61906185), Youth Innovation Promotion Association of CAS China (No. 2020357), Shenzhen Science and Technology Innovation Program (Grant No. KQTD20190929172835662), Shenzhen Basic Research Foundation (No. JCYJ20210324115614039 and No. JCYJ20200109113441941). This work was supported by Alibaba Group through Alibaba Innovative Research Program.

\bibliography{ref}
\bibliographystyle{acl_natbib}

\appendix

\section{More Implementation Details}
\label{sec:more-implementation-details}
In the pre-training, \name is initialized with \electra \cite{clark2020electra}. Similar to ELECTRA which is consist of a generator $\mathcal{G}$ and a discriminator $\mathcal{D}$, we also employ the replaced token detection objective to further improve the text-to-SQL pre-training. Concretely, given the input $I_t$ $\label{eq:input}$ (defined in Eq. (1)) of the $t$-th turn, the generator with masked language modeling (MLM) selects a random set of positions and replaces these positions with \texttt{[MASK]}, and then learns to predict the original tokens of the masked-out tokens. The chance of each token being masked out is 15\%. We denote the loss function of the generator as $\mathcal{L}_{\rm MLM}$. In addition, we also train the discriminator to predict whether the each token is the same as the original token. We denote the loss function for training the discriminator as $\mathcal{L}_{\rm Dis}$. Finally, we combine the loss functions of the generator $\mathcal{G}$ and the discriminator $\mathcal{D}$ to form the overall objective function for replaced token detection (RTD) as:
\begin{equation}
    \mathcal{L}_{\rm overall} = \mathcal{L}_{\rm Dis} + \gamma\mathcal{L}_{\rm MLM}
\end{equation}
We refer the readers to \cite{clark2020electra} for the implementation details of the RTD objective. $\gamma$ is a hyperparameter controlling the impact of $\mathcal{L}_{\rm MLM}$. In this work, the impact factor $\gamma$ is set to 5.
Our codebase is built on huggingface library \cite{wolf2019huggingface}.

We use LGESQL as our downstream model. For a fair comparison, all LGESQL experiments are trained for 100 epoch. The learning rate is 1e-4 and weight decay is 0.1. And we adopt a more carefully optimization for our \name encoder with layer-wise learning rate decay coefficient 0.8. Batch size is 10 and the maximum gradient norm is 5. Other hyperparameters are the same as in \cite{cao2021lgesql}.

\section{Details of \sparc and \cosql}
\label{sec:data-details}
We evaluate the effectiveness of \name on two context-dependent text-to-SQL parsing benchmarks: \sparc \cite{yu2019sparc} and \cosql \cite{yu2019cosql}. Concretely, \sparc is a collection of cross-domain context-dependent dataset, which consists of about 4.3k  question sequences and 12k+ individual questions annotated with SQL queries. \cosql is a conversational text-to-SQL corpus, which contains about 3k dialogues and 10k+ annotated SQL queries.  Both \sparc and \cosql query 200
complex databases spanning across 138 domains.
Table \ref{tab:dataset_stats} reports the statistics  of \sparc and \cosql datasets in detail.

\section{Generalization of \name}
\label{sec:generalization-star}
We also evaluate the generalization of our pre-training objectives by using \roberta as our initialization model, rather than applying ELECTRA. The experimental results are shown in Table \ref{init-model}. In a similar trend, \name that is 
initialized with \roberta performs significantly better than the original \roberta, which to some extent verifies the generalization of the proposed pre-training objectives, no matter what initialization models are used to train \name.

\begin{table}
\begin{center}
\scalebox{0.9}{
\begin{tabular}{lcc}
\toprule
 & \textbf{\sparc} & \textbf{\cosql}\\ 
\midrule
\# Question Sequences & $4,298$ & $3,007$\\
\# Train & $3,034$ & $2,164$\\
\# Dev & $422$ & $293$\\
\# Test & $842$ & $551$ \\
\# User Questions & $12,726$ & $15,598$ \\
\# Databases & $200$ & $200$ \\
\# Domain & $138$ & $138$ \\
Avg.len & $8.1$ & $11.2$ \\
Vocab & $3,794$ & $9,585$ \\
System Response & no & yes\\
\bottomrule
\end{tabular}
}
\end{center}
\caption{Details of SParC and CoSQL Dataset.}
\label{tab:dataset_stats}
\end{table}

\begin{table}
	\centering
	\scalebox{0.8}{
	\begin{tabular}{l|cc|cc}
		\toprule 
		{\multirow{3}*{\textbf{Model}}} & \multicolumn{2}{c|}{\sparc} & \multicolumn{2}{c}{\cosql} \\
		\cmidrule{2-5}
		\multirow{2}{*}{} &
		\multicolumn{1}{c}{QM} &
		\multicolumn{1}{c|}{IM} &
		\multicolumn{1}{c}{QM} &
		\multicolumn{1}{c}{IM} \\
		\midrule
		\roberta & 61.6 & 41.2 & 51.9 &  20.8\\
		\name (init. with \roberta) & \textbf{65.0} & \textbf{45.1} & \textbf{54.1} & \textbf{25.3}  \\
		\bottomrule
	\end{tabular}
	}
	\caption{Results of \name which is initialized with \roberta on the dev sets of both \sparc and \cosql.}
	\label{init-model}
\end{table}

\begin{figure*}
    \centering
    \includegraphics[width = 16cm]{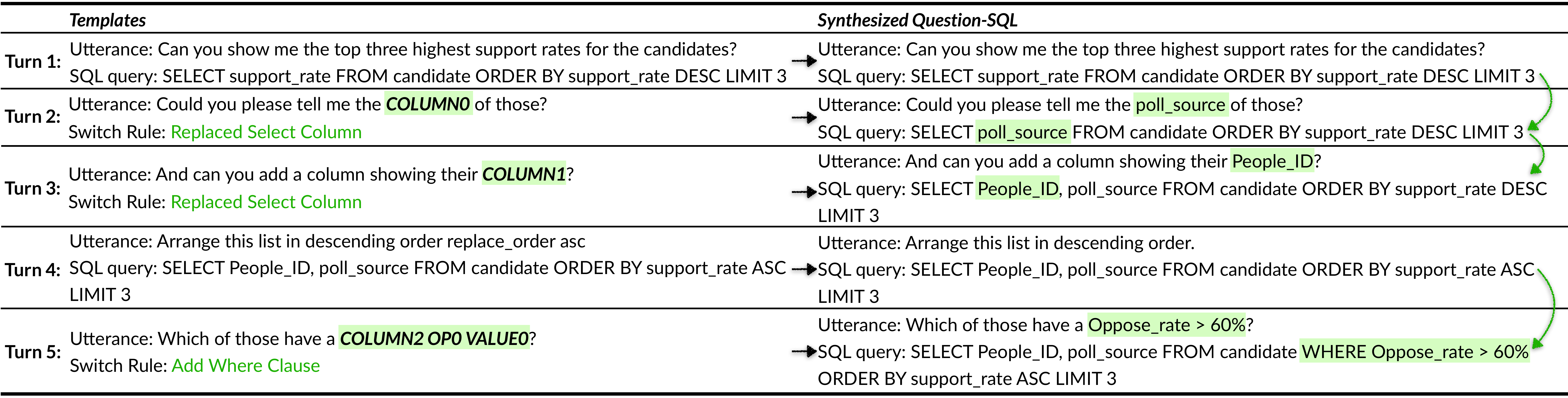}
    \caption{An example of synthetic text-to-SQL conversation.}
    \label{fig:synthetic-example}
\end{figure*}

\section{Details of Data Construction}
\label{sec:data-constraction}
In this paper, we synthesize a new large-scale pre-training dataset which consists of about 480K high-quality context-dependent text-to-SQL conversations. Specifically, we first generate single-turn question-SQL pairs by exploiting the \spider, \sparc and \cosql datasets.

\subsection{Single-turn Question-SQL Pairs}
\label{single-turn-data}
To obtain sufficient high-quality single-turn question-SQL pairs, we carefully examine currently available sources and generate question-SQL pairs from \spider, \sparc and \cosql datasets. Specifically, we collect the original single-turn question-SQL pairs from the dataset \spider which is one of the largest single-turn cross-domain text-to-SQL corpora. For the context-dependent text-to-SQL datasets \sparc and \cosql, we generate a new question for each SQL query instead of using the original NL questions since they may contain ellipsis and anaphora that refers to earlier items in the conversations, resulting in low-quality question-SQL pairs. In particular, we employ the SNOWBALL framework \cite{shu2021logic} with BART to generate the question based on each SQL query, which employs an iterative training procedure by recursively augmenting the training set with quality control. 

\subsection{Context-dependent Text-to-SQL Conversations}
To expand the single-turn question-SQL pairs to context-dependent text-to-SQL conversations, we first convert SQL queries into their structured formats. For example, we convert the SQL query ``\textit{SELECT support\_rate FROM candidate ORDER\_BY support\_rate LIMIT 3}'' into a set of SQL states as \{\texttt{SELECT}: \texttt{[support\_rate]}, \texttt{FROM}: \texttt{[candidate]}, \texttt{ORDER\_BY}: \texttt{[support\_rate]}, other SQL keywords: \texttt{[NONE]}\}. 

Then, following \cite{yu2020score}, we study 600 examples from the training set of both SPARC and CoSQL datasets, and induce about 100 follow-up question-grammar templates. Each template consists of a pair of (i) a context-free question template (e.g., ``\textit{Could you please tell me the \texttt{[COLUMN0]} of those?}'') where the typed slot \texttt{[COLUMN0]} represents the mention of schema, and (ii) its corresponding operation grammar (e.g., ``\textit{replaced select column}'') that contains context switch labels of the question templates. 

Finally, for a single-turn question-SQL pair constructed in Section \ref{single-turn-data} with database $d$, we randomly choose a created question-grammar template. We sample the values for typed slots in the template and get the synthesized NL question as well as its corresponding SQL query if the previous SQL query satisfies the constraints in the sampled template (e.g., the SQL query contains the mentioned schema); otherwise, another question-grammar temple is sampled until we successfully synthesize the next question-SQL pair. 
Then, we consider the synthesized question-SQL pair as a new start and repeat the above process until we obtain the context-dependent text-to-SQL conversation consisting of $T$ turns of question-SQL pairs. Figure \ref{fig:synthetic-example} shows an example of synthetic text-to-SQL conversation with five turns. 

\end{document}